# Can we learn where people go?

**Marion Gödel[1,2], Gerta Köster[1], Daniel Lehmberg[1,2], Manfred Gruber[1], Angelika Kneidl[3], Florian Sesser [3]**

[1]Munich University of Applied Sciences
Lothstrasse 64, 80335 Munich, Germany
Marion.Goedel@hm.edu, Gerta.Koester@hm.edu, Daniel.Lehmberg@hm.edu, Manfred.Gruber@lrz.fh-muenchen.de
[2]Technical University of Munich,
Boltzmannstrasse 6, Munich, Germany
[3]accu:rate GmbH Institute for crowd simulation
Rosental 5, 80331 Munich, Germany
ak@accu-rate.de, fs@accu-rate.de

***Abstract*** *– In most agent-based simulators, pedestrians navigate from origins to destinations. Consequently, destinations are essential input parameters to the simulation. While many other relevant parameters as positions, speeds and densities can be obtained from sensors, like cameras, destinations cannot be observed directly. Our research question is: Can we obtain this information from video data using machine learning methods? We use density heatmaps, which indicate the pedestrian density within a given camera cutout, as input to predict the destination distributions. For our proof of concept, we train a Random Forest predictor on an exemplary data set generated with the Vadere microscopic simulator. The scenario is a crossroad where pedestrians can head left, straight or right. In addition, we gain first insights on suitable placement of the camera. The results motivate an in-depth analysis of the methodology.*

***Keywords*: (4-8)** pedestrian dynamics, predictive simulation, machine learning, random forest

## 1. Introduction

It is a shared goal of crowd simulation experts to look into the future for at least a few minutes to predict dangers like extremely high densities that might evolve. State-of-the-art microscopic models are, in principle, capable of producing correct crowd flows in many relevant situations, provided they get correct input parameters. The basic idea of predictive crowd analysis is to gather these input parameters online from sensors. Many relevant parameters, like positions, speeds and densities can be obtained from cameras, even if the speed and accuracy with which the data is acquired may be insufficient for prediction at the moment. However, some essential input parameters cannot be observed directly: chief among them are destinations where people go.

Agent-based microscopic crowd simulations use destinations to navigate pedestrians. This holds especially for all simulations based on a floor field. The floor field indicates for each position of the scenario the distance to the destination(s). As a consequence, destinations are a crucial input parameter for the simulation. Nevertheless, in reality destinations are often unknown. They need to be chosen according to experience or statistics. In addition, a previous study has shown that the destinations of pedestrians have a high impact on the simulation output [1].

This work is performed in the context of the S[2]UCRE research project (www.s2ucre.de). The goal of the project is to set up a control cycle that performs short-term predictions which are continuously compared to the latest video footage. Every time the simulation is started, a destination needs to be assigned to each pedestrian.

To our knowledge, currently only very few publications are available on retrieving information about pedestrian's destinations [2]. In practical applications of pedestrian simulations, there are several ways to assign the destinations: First, for simulations of evacuation scenarios, known gathering points, emergency exits or simply the closest exit can be chosen. Second, for festivals or events as well as museums and infrastructural buildings, destinations can be assigned based on experience of the organizer, timetables or visitor surveys. Unfortunately, none of these approaches is reliable enough to establish quantitative alignment within a control cycle.

Pedestrian density heatmaps will be available at a later stage of the S$^2$UCRE project as an input to the simulation. In addition, methods of density estimation, the derivation of heatmaps from video footage is expected to become state-of-the-art at some point [3]. That is why we use heatmaps as a basis. Our goal is to extract information about pedestrians' destinations from such heatmaps deploying machine learning techniques. In fact, the research question is: Is there enough information in the heatmaps to predict the distribution of destinations?

Machine learning has become very popular over the last years. There are first approaches to apply machine learning models in the context of pedestrian dynamics. The applications range from crowd counting and density estimation [3] to the prediction of pedestrian movement [4]. Furthermore, destination prediction aims to predict the destinations of people of a certain audience based on trajectories and prior knowledge of possible destinations [5,6]. In our case, the destinations are predicted solely based on a dataset of density heatmaps and our knowledge about possible destinations. We do not utilize user-dependent data.

Our solution proposal for destination prediction is data-driven. That means deploying information from videos or other sources for simulation. This ansatz has already been employed for traffic dynamics. There are two main approaches of data-driven modelling: First, deriving a model from a data set instead of using an equation or rule-based model. Second, complementing existing rule-based models with parameters derived from videos. For the former, [7-11] are of particular interest in our context. The latter was carried out in [12] together with the Social Force Model. In our application, we want to predict macroscopic traffic quantities, such as density and flow, through simulations with an explanatory microscopic model, the Optimal Steps Model [13,14]. The simulation is complemented with parameter learning for the destination distributions from video footage.

For a proof of concept, we will use heatmaps generated from simulations using the simulation framework Vadere [15] instead of actual video footage. We focus on scenarios that appear relevant in the context of the research project S$^2$UCRE. More specifically, we analyse a crossroad in a pedestrian area. A Random Forest predictor is trained on the snapshots to predict decision distributions at the crossroad.

## 2. Methods and Configuration

Our simulations are performed using the Optimal Steps Model within the Vadere framework. The choice of the next step consists of three parts: First, for each destination a floor field which indicates the distance towards the destination is calculated. Second, at every time step and for each pedestrian repulsion from obstacles and other pedestrians is added to the static floor field. Thus, the ensuing floor field codes negative utility, or cost, of a position. Third, agents find the best next position with respect to the floor field within a circle of their step lengths. A detailed description can be found in [14]. This procedure for choosing the next step explains why targets are a crucial input parameter.

We chose Random Forest [16] as machine learning algorithm to predict the destination distribution based on the density heatmap. A Random Forest is a collection of decision trees whose results are aggregated into one final result. Random Forest is known to be a robust algorithm and easy to apply as there are only few tuning parameters [16,17]. Furthermore, it can be used directly for high-dimensional problems. In addition, Random Forest often works well without heavy tuning of parameters [18].

This study is conducted using the Python implementation provided by scikit-learn [19]. Computations are performed on a platform with an Intel(R) Xeon X5672 CPU with 3.20GHz 4 Cores and 16 GB DDR3 RAM.

In the following, we describe our configuration to adapt the methods for the chosen problem.

### 2.1. Scenario

A simple, symmetric scenario is chosen for the proof of concept. See Fig. 1. The pedestrians walk from an origin to one of three destinations (left / straight /right). The camera cutout used as baseline for the heatmap generation is shown in red. For an arbitrary time step, the positions of the pedestrians are indicated in blue. In addition, the trajectories for all pedestrians in the scenario at the chosen time step are shown. The scenario only serves to simulate the trajectories, which are used to generate the heatmaps and derive the corresponding destination distribution.

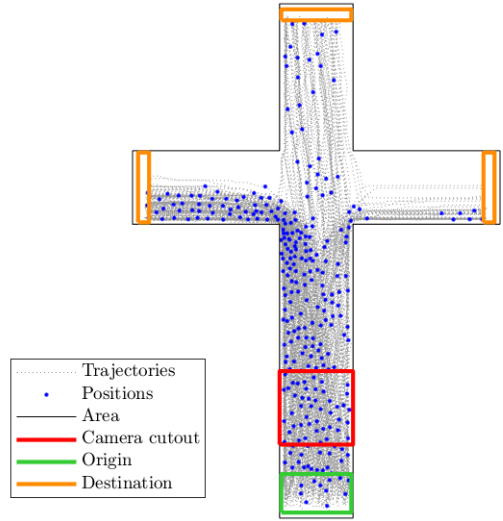

Fig. 1: Scenario for the proof of concept. Pedestrians (blue) walk from an origin (green) to one of three destinations (orange). The camera cutout is indicated in red.

### 2.2. Density heatmaps

The heatmaps are generated from a set of positions for each time step for a defined camera cutout. At an arbitrary time step, there are n pedestrians at positions $x_i$ located within the cutout. The Gaussian density at position $z$ is calculated as

$$D_p(z) = \frac{d_p^2 \sqrt{3}}{4\pi S^2} \sum_{i=1}^{n} \exp\left(-\frac{\| x_i - z \|^2}{2S^2}\right) \qquad (1)$$

with torso diameter $d_p = 0.195$ m for each pedestrian and scale factor $S = 0.7$ [13,20]. In Fig. 2 a set of five consecutive density heatmaps is shown.

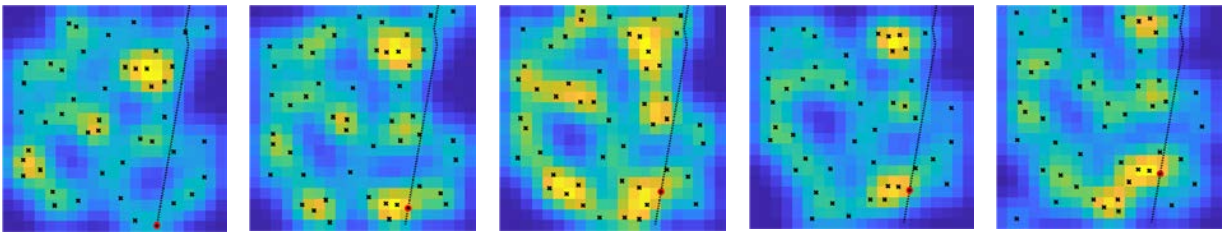

Fig. 2: Five consecutive density heatmaps for a camera cutout of 10x10 m with a resolution of 0.5 m. Pedestrian positions are indicated as black crosses. For one exemplary pedestrian the trajectory is shown as dotted line.

## 2.3. Configuration of Random Forest

The actual density values describing the two-dimensional heatmaps are rearranged to a vector (row-wise). The combination of one vector and the corresponding destination distribution forms a sample for the Random Forest model. As a consequence, each pixel value in the heatmap serves as a feature. Table 1 shows the response corresponding to the samples visualized in Fig. 2. The response are the percentages of pedestrians within the cutout heading in the three directions. One can see that the heatmaps differ significantly even though the distribution of the pedestrians on the destinations varies only slightly. As a result, one cannot simply derive the destination distribution by looking at a heatmap.

Table 1: Response for Random Forest: Actual percentage of pedestrians heading left (L), straight (S) and right (R).

| Percentage of pedestrians heading | | | Percentage of pedestrians heading | | | Percentage of pedestrians heading | | | Percentage of pedestrians heading | | | Percentage of pedestrians heading | | |
|---|---|---|---|---|---|---|---|---|---|---|---|---|---|---|
| L | S | R | L | S | R | L | S | R | L | S | R | L | S | R |
| 28.3 | 39.1 | 32.6 | 28.0 | 40.0 | 32.0 | 32.0 | 38.0 | 30.0 | 32.6 | 41.3 | 26.1 | 32.0 | 40.0 | 28.0 |

The problem is posed as a multi-dimensional regression problem. The number of response dimensions is the number of identified destinations. In the considered crossroad scenario, there are three destinations. The advantage of this configuration compared to a classification problem is twofold: First, the responses are not a finite set of classes but continuous between zero and one. Consequently, one would have to design classes that are small enough for the application and large enough to contain enough samples in order to use classification. Second, applying a regression, Random Forest can predict destination distributions that were not part of the training set. The drawback of a multi-dimensional response configuration is that the destination distributions do not necessarily add up to 100%. We overcome this problem by normalizing the model predictions.

## 2.4. Performance measure

The Random Forest routines can provide an out of bag error estimate as well as the coefficient of determination. However, both quantities are applied before the model predictions are normalized. Therefore, we split our dataset into a training and a test set and compare the prediction on the test set to the corresponding response to evaluate the model. In a first step, we evaluate the Euclidean norm of the difference vector. In the second step, we derive a relative error. Since we normalize the prediction, the maximum error

$$e_{max} = \sqrt{2 \cdot 100^2} \approx 141.42 \ . \tag{2}$$

occurs if we predict that all pedestrians within the cutout head left while, in fact, they head right (or straight): The relative error is then

$$e = \frac{y - \hat{y}}{e\_max} \cdot 100 \ , \tag{3}$$

where $y$ is the response on the test set and $\hat{y}$ is the prediction on the test set. Thereby we obtain a relative error of the prediction that is easy to interpret. Thus, the relative error is used as a measure.

## 3. Results and Discussion

We predict the percentage of pedestrians heading in three directions: left, straight and right. The scenario used for the simulation is shown in Fig. 1. The simulation serves as a basis for the heatmap generation. All results are obtained using the same data set of simulations described in the following. The heatmaps are generated from 50 simulation runs, each of 500 seconds, performed with Vadere. The first 12 seconds are cut off in order to let the system settle and to observe a well filled camera cutout. In total, we obtain 3050 heatmaps. The available dataset is split into a training set (80%) and a test set (20%).

Each pedestrian in the simulation is randomly assigned one of the three destinations according to the current destination distribution. The destination distribution is altered with the generation of every $100^{th}$ pedestrian. Thus, we make sure that a variety of destination distributions is considered. For each direction, we train one separate Random Forest model. In principal, Random Forest can handle multi-variable responses, but, with separate models, we observe an improvement over using one model that predicts all directions.

The goal of this paper is to offer a proof of concept that one can indeed extract information about destinations from the density heatmaps. As preparation for practical application on real video data, we vary the size and position of the camera cutout and analyse the impact on the accuracy and on computation time.

### 3.1. Proof of concept

The camera cutout is placed as shown in Fig. 1. Its dimensions are 10 meters $\times$ 10 meters. The resolution of the heatmaps is set to 0.5 meters. Consequently, the heatmaps consist of $20 \times 20$ pixels. The correlation between the heatmaps is minimized by choosing a framerate corresponding to the time that pedestrians need to pass the camera cutout. Since the camera cutout is of length 10 meters and the mean free-flow velocity is 1.34 m/s, pedestrians need roughly 7.46 seconds to cross it. Thus, we choose a frame rate of 0.05 Hz, that is, one heatmap every eight seconds. Therefore, a pedestrian is typically considered in only one heatmap.

Fig. 3 shows the performance and computation time in dependency of the number of trees per model. The computation time increases linearly with increasing number of trees while the Euclidean error decreases up to roughly 20 trees. As a result, we use 20 trees for the proof of concept. We perform the split of training and test set, as well as training and test of the Random Forest predictor five times. We obtain a mean Euclidean error of 12.24% and standard deviation 6.93%. Thus, the method is able to predict where people are heading at the crossing with an accuracy of roughly 88%. The average computation time to train the forests is 4.28 seconds.

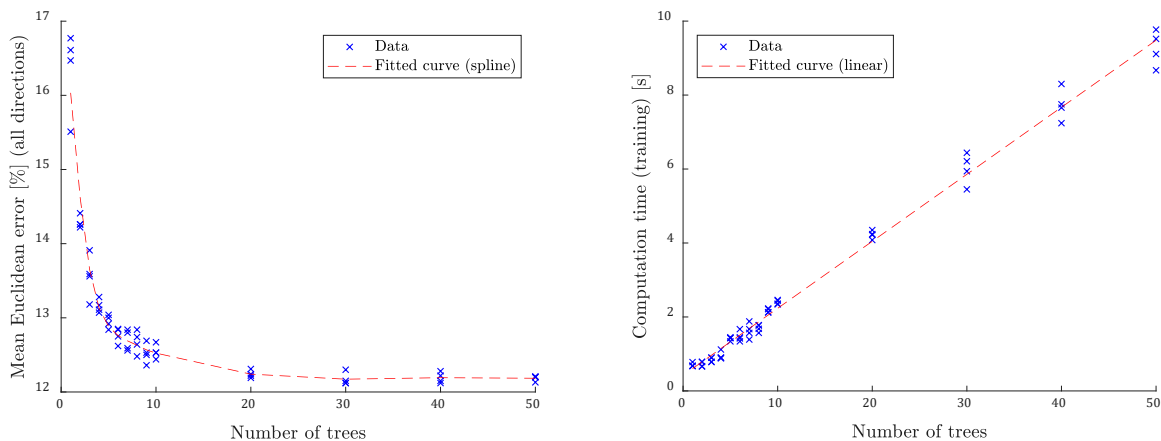

(a) Mean Euclidean error of the prediction on the test set.

(b) Computation time necessary to train the forest.

Fig. 3: Results of Random Forest for different numbers of trees per forest.

## 3.2. Placement of the camera cutout

Since the proof of concept was successful, we take a closer look at the parameters. One vital parameter for the application of the method on real video footage is camera placement. First, we vary the distance of the camera cutout to the crossing and analyse the impact on the quality of the prediction. Figure 4 (a) shows the varied positions of the camera cutout. In the second step, we analyse the impact of the size of the camera cutout on the predictions. We limit ourselves to cutouts that cover the whole width of the street. Fig. 4 (b) shows eight sizes used for the analysis.

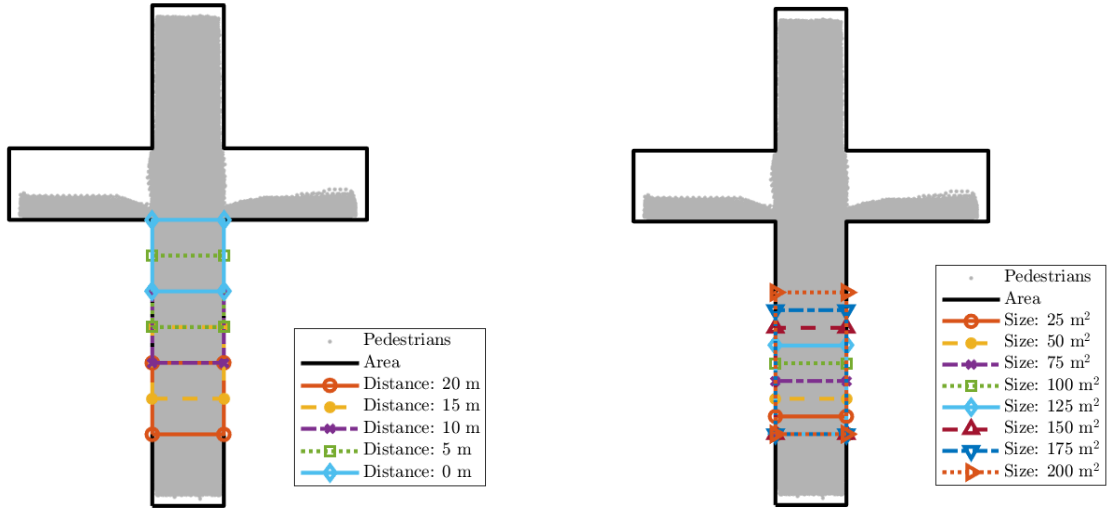

(a) Position of the camera cutout in the corridor.  (b) Size of the camera cutout.

Fig. 4: Variation of the camera cutout.

Fig. 5 (a) shows the mean Euclidean error in dependency of the position of the camera cutout. The positions are defined by the distance of the upper edge of the cutout to the lower position of the crossing. We observe that the quality of the prediction only slightly increases when placing the camera cutout closer to the crossing. Thus, the camera does not have to be placed directly in front of the crossing for a sound prediction. That is a big advantage for the practical application, since camera positions are often determined by the space where equipment can be mounted. Also, monitoring an area further away from the crossing, gives us the necessary time frame for the prediction.

In Fig. 5 (b) the results for different sizes of the camera cutouts are depicted. Since the number of features increases with the size of the cutout, the computation time increases as well. We observe a linear increase. The results reveal that the quality of the prediction is highly dependent on the size of the camera cutout. Therefore, in this scenario, the camera cutout should cover at least 75 m². This size yields a mean Euclidean error of 13.15%. Nevertheless, one has to keep in mind that these results are based on simulated heatmaps not on actual video footage.

Besides the insights that we have gained on the impact of the placement of the camera, both variations also serve as plausibility checks. As expected, the quality of the prediction increases both with placing the camera cutout closer to the crossing and with increasing the size of the camera cutout. In combination with alignment between two independent implementations, this verifies our code.

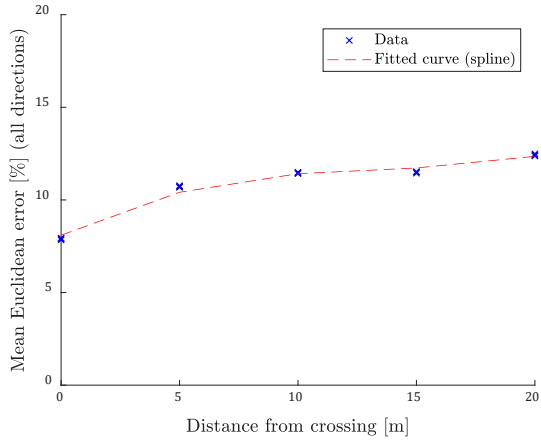 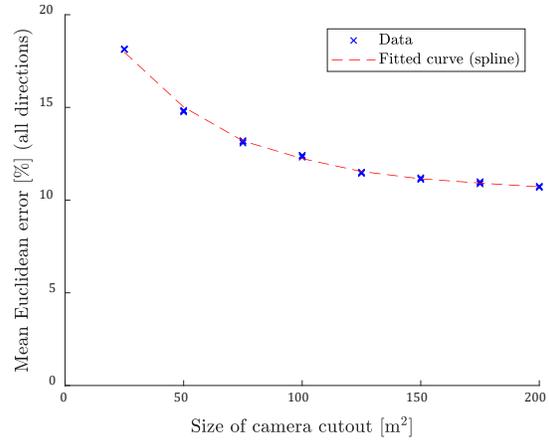

(a) Different positions of the camera cutouts.  (b) Different sizes of the camera cutout.

Fig. 5: Performance of Random Forest for different camera cutouts. The position represented by the distance of the upper edge of the cutout to the lower position of the crossing. For each position and for each size, five runs were performed.

## 4. Conclusion and Outlook

In this publication, we deliver a proof of concept for configuring a machine learning method to obtain the destinations of pedestrians based on density heatmaps. It was carried out based on a data set of pedestrians' trajectories generated with our simulation framework Vadere. The chosen method (Random Forest) is able to predict the distribution of pedestrians on destinations with an accuracy of up to 90%. In addition, we performed plausibility checks by modifying the size and position of the camera cutout. Coincidently, we gained insight on the impact of the camera placement: While the position of the camera cutout within the street seems to have little effect, the size of the camera cutout has a significant influence on the quality of the prediction. Consequently, the camera cutout should be chosen as large as possible. In both cases, the camera cutout was selected such that the whole width of the street is covered.

These first results show that the methodology has great potential. We plan to further investigate the approach. In the next step, one could adapt the method to different scenarios to analyse the robustness of the technique. In addition, one needs to study the effect of parameters such as the number of pedestrians in the scenario. Also, we did not exploit the time dependency of adjoining snapshots in this contribution. We expect the prediction quality to profit from adding a time component to the input data. One approach is to use a group of heatmaps instead of a single heatmap. Furthermore, while Random Forest has proven to be robust in this scenario, there are many other machine-learning algorithms. In particular, Convolutional Neural Networks are able to process images as the density heatmaps directly as input and might be able to exploit the topological information. Finally, the main goal is to adapt the methodology for the application on real video footage.

## Acknowledgements

The authors would like to thank the students who have carried out first investigations within a project study at Munich University of Applied Sciences: Julian Bauer, Rebecca Brydon, Patrick Gabler, Katja Gruenewald, Lisa-Marie Grundmann, Hubert Hager, Sebastian Klohn, Carsten Kruse, Tim Lauster, Julia Maier, Sarah Nesner, Do Nhu Nguyen, Luca Spataro, Anita Steinberger, Veronika Zwickenpflug. This work was funded by the German Federal Ministry of Education and Research through the project S2UCRE (grant number 13N14464). The authors acknowledge the support by the Faculty Graduate Center CeDoSIA of TUM Graduate School at Technical University of Munich and the research office FORWIN at Munich University of Applied Sciences.

## References

[1] M. Davidich and G. Köster, "Predicting Pedestrian Flow: A Methodology and a Proof of Concept Based on Real-Life Data," *PLoS ONE*, vol. 8, no. 12, pp. 1-11, 2003.

[2] P. M. Kielar, A. Borrmann, "Modeling pedestrians' interest in locations: A concept to improve simulations of pedestrian destination choice," *Simulation Modelling Practice and Theory*, vol. 61, pp. 47-62, 2016.

[3] V. A. Sindagi and V.M. Patel, "A Survey of Recent Advances in CNN-based Single Image Crowd Counting and Density Estimation," *Pattern Recognition Letters*, vol. 107, pp. 3-16, 2018.

[4] Y. Ma, E. W. M. Lee and R. K. K. Yuen, "An Artificial Intelligence-Based Approach for Simulating Pedestrian Movement," in *IEEE Transactions on Intelligent Transportation Systems*, vol. 17, no. 11, pp. 3159-3170, 2016.

[5] J. Krumm and e. Horvitz, "Predestination: Inferring destinations from partial trajectories," in *Proceedings of the 8th International Conference on Ubiquitous Computing UbiComp '06,* pp. 243-260, 2006.

[6] J. K. Laurila et al., "The mobile data challenge: Big data for mobile computing research," in *Proceedings of the Workshop on the Nokia Mobile Data Challenge, in Conjunction with the 10th International Conference on Pervasive Computing,* p. 1-8, 2012.

[7] A. Bera, S. Kim, D. Manocha, "Online parameter learning for data-driven crowd simulation and content generation," *Computers and Graphics (Pergamon)*, vol. 55, pp. 68-79, 2016.

[8] N. Bisagno, N. Conci, B. Zhang, "Data-Driven crowd simulation", in *14th IEEE International Conference on Advanced Video and Signal Based Surveillance*, Lecce, Italy, 2017, pp.

[9] F. Dietrich, "Data-Driven Surrogate Models for Dynamical Systems", Ph. D. dissertation, Dept. Scientific Computing, Technical University of Munich, Germany.

[10] S. Kim et al., "Interactive and adaptive data-driven crowd simulation," in *Proceedings of IEEE Virtual Reality*, Greenville, SC, 2016, pp. 29-38.

[11] J. Porzycki et al., "Dynamic data-driven simulation of pedestrian movement with automatic validation," in *Traffic and Granular Flow '13*, Jülich, Germany, 2013, pp. 129-136.

[12] B. Liu et al., "A social force evacuation model driven by video data*,"* *Simulation Modelling Practice and Theory*, vol. 84, pp. 190-203, 2018.

[13] M. J. Seitz and G. Köster, "Natural discretization of pedestrian movement in continuous space," *Physical Review E*, vol. 86, no. 4, 2012.

[14] I. von Sivers and G. Köster, "Dynamic Stride Length Adaptation According to Utility and Personal Space", *Transportation Research Part B: Methodological*, vol. 75, pp. 104-117, 2015.

[15] Vadere Crowd Simulation (2016) [Online]. Available: https://gitlab.lrz.de/vadere/vadere

[16] L. Breiman, "Random Forests," *Machine Learning*, vol. 45, no. 1, pp. 5-32, 2001.

[17] A. Cutler, D.R. Cutler, J.R. Stevens, "Random Forests" in *Ensemble Machine Learning*, C. Zhang, Y. Ma, Ed. Boston, 2012, pp. 157-175.

[18] A. C. Müller and S. Guido, „Introduction to Machine Learning with Python – A Guide for Data Scientists", O'Reilly Media, Inc., 2017, p.89.

[19] F. Pedregosa et al., "Scikit-learn: Machine Learning in Python*,"* *Journal of Machine Learning Research*, vol. 12, pp. 2825-2830, 2011.

[20] D. Helbing, A. Johansson, H. Z. Al-Abideen, "Dynamics of crowd disasters: An empirical study," *Physical Review E*, vol. 75, no. 4, 2007.